\documentclass[conference]{IEEEtran}
\IEEEoverridecommandlockouts
\usepackage{cite}
\usepackage{amsmath,amssymb,amsfonts}
\usepackage{algorithmic}
\usepackage{graphicx}
\usepackage{textcomp}
\usepackage{xcolor}
\usepackage{url}

\usepackage{listings}
\usepackage{color}

\definecolor{dkgreen}{rgb}{0,0.6,0}
\definecolor{gray}{rgb}{0.5,0.5,0.5}
\definecolor{mauve}{rgb}{0.58,0,0.82}

\def\BibTeX{{\rm B\kern-.05em{\sc i\kern-.025em b}\kern-.08em
    T\kern-.1667em\lower.7ex\hbox{E}\kern-.125emX}}
\begin{document}

\title{Improving the Generation of VAEs with High Dimensional Latent Spaces by the use of Hyperspherical Coordinates\\
\thanks{This work was partially supported by the CSIRO AI4M Research Program.}
}

\author{\IEEEauthorblockN{1\textsuperscript{st} Alejandro Ascarate}
\IEEEauthorblockA{\textit{School of Electrical Engineering and Robotics} \\
\textit{Queensland University of Technology}\\
Brisbane, Australia \\
a.ascaratecastro@hdr.qut.edu.au}
\and
\IEEEauthorblockN{2\textsuperscript{nd}  Leo Lebrat}
\IEEEauthorblockA{\textit{School of Electrical Engineering and Robotics} \\
\textit{Queensland University of Technology}\\
Brisbane, Australia \\
leo.lebrat@qut.edu.au}
\and
\IEEEauthorblockN{3\textsuperscript{rd} Rodrigo Santa Cruz}
\IEEEauthorblockA{\textit{Data61} \\
\textit{CSIRO}\\
Brisbane, Australia \\
rodrigo.santacruz@csiro.au}
\and
\IEEEauthorblockN{4\textsuperscript{th} Clinton Fookes}
\IEEEauthorblockA{\textit{School of Electrical Engineering and Robotics} \\
\textit{Queensland University of Technology}\\
Brisbane, Australia \\
c.fookes@qut.edu.au}
\and
\IEEEauthorblockN{5\textsuperscript{th} Olivier Salvado}
\IEEEauthorblockA{\textit{School of Electrical Engineering and Robotics} \\
\textit{Queensland University of Technology}\\
Brisbane, Australia \\
olivier.salvado@qut.edu.au}
}

\maketitle

\begin{abstract}
Variational autoencoders (VAE) encode data into lower-dimensional latent vectors before decoding those vectors back to data. Once trained, decoding a random latent vector from the prior usually does not produce meaningful data, at least when the latent space has more than a dozen dimensions. In this paper, we investigate this issue by drawing insight from high dimensional statistics: in these regimes, the latent vectors of a standard VAE are by construction distributed uniformly on a hypersphere. We propose to formulate the latent variables of a VAE using hyperspherical coordinates, which allows compressing the latent vectors towards an island on the hypersphere, thereby reducing the latent sparsity and we show that this improves the generation ability of the VAE. We propose a new parameterization of the latent space with limited computational overhead.
\end{abstract}


\section{Introduction}

In today's machine learning landscape, and deep learning in particular, one of the main mathematical tools to represent data (e.g. images) are high dimensional (HD) Euclidean spaces.

However, our intuition about Euclidean geometry stems from the physical world and everyday life, which are low dimensional spaces (mostly two and three dimensions). This presents a challenge because HD spaces behave, mathematically speaking, in very different ways than their low dimensional counterparts, often in ways that seem counterintuitive or even paradoxical if interpreted through low dimensional intuition. 

We will highlight how issues associated with VAE are related to hypervolume and the creation of voids and sparsity in latent spaces, hampering their performance to generate meaningful new samples, even when reconstruction metrics can be optimized very well.

Finally, we will propose a novel method that parametrize latent data on the hypersphere with hyperspherical coordinates. This allows manipulating the data distribution on the latent manifold more effectively. In particular, we use it to compress the latent manifold volume and reduce the sparsity. This is made possible thanks to an efficient transformation between Cartesian and hyperspherical coordinates, which can be implemented with minimal computational overhead using a fully vectorized algorithm, for high enough dimensions (it becomes costly in the very large case; but, as our results show, those cases are of no practical interest from the point of view of the metrics we are checking, see Fig.\ref{fig:2}). 

\subsection{Variational Autoencoder}

An autoencoder (AE) is a common self-supervised method to encode the data into a latent space of lower dimension $n$. Variational autoencoder (VAE) \cite{DiederikPKingma2014Auto-EncodingBayes, Kingma2019AnAutoencoders} introduces a prior to control the distribution of the latent, allowing to sample the latent space and untangle the dimensions \cite{Rolinek2019VariationalAccident, Bhowal2024WhyDisentanglement}. In its simplest implementation, a VAE consists of a probabilistic encoder which, for each input point $x \in X$ from a dataset $X$, produces a latent distribution \(q_{x}=\mathcal{N}(\mu_{x},\sigma_{x})\). During training, the reparameterization trick is used: the encoder estimates $\mu$ and $\sigma$, and a sample $z$ from $q_{x}$ is computed as $z = \mu + \epsilon \odot \sigma$, where $\odot$ denotes element-wise multiplication. Then, the decoder is applied to this sample to obtain the reconstructed datapoint, $\hat{x}_z$. The VAE's loss can then be interpreted as an AE (with its Mean Square Error, MSE, loss) with an additional term, $\text{KLD}\left(\mathcal{N}(\mu_{x},\sigma_{x})\parallel\mathcal{N}(0,I)\right)$, that regularizes the latent space by forcing each of the encoded distributions to become similar to a prior one ($\mathcal{N}(0,I)$ in this implementation), where \textit{KLD} refers to the Kullback-Leibler divergence. The two mentioned terms are computed over a mini-batch of size $N_b$:

\begin{align}
\text{MSE}(x,\hat{x}_z) &= \frac{1}{N_b} \sum^{N_b}_{l=1}\parallel x_l-\hat{x}^l_z\parallel^{2},\\
\text{KLD}(z,\epsilon) &= -\frac{1}{2} \sum^{N_b}_{l=1} \sum^{n}_{k=1} \left(  1+\log\,(\sigma^2_{k,l})-\mu^2_{k,l} - \sigma^2_{k,l} \right).
\end{align}

The final cost to be optimized weights the two terms with the gain $\beta$ \cite{Higgins2017-VAE:FRAMEWORK}:

\begin{equation}
\mathcal{L} = \text{MSE}(x,\hat{x}_z) + \beta \text{KLD}(z,\epsilon).
\label{eq:VAEcost}
\end{equation}

The prior, and thus the latent, is a high dimensional independent multivariate Gaussian, which has specific properties that we briefly recall in the next section, by way of background for the following sections. 

\subsection{High Dimensional Spaces}

A multivariate Gaussian sampling in a HD Euclidean space of dimension \(n\) is such the probability to find samples close to the origin is close to zero (despite having maximum probability density) and most of the samples lie close to a \((n-1)-\)hypersphere, \(\mathbb{S}_{\sqrt{n}}^{n-1}\), of radius $\sqrt{n}$. The distribution of the norm of those samples follows a $\chi(n)$ distribution. Therefore, the samples are located within a region very close to the hypersphere, region which becomes very thin in high dimensions, relative to the radius $\sqrt{n}$. These effects are called `concentration of measure', in the mathematical literature.

As \(n\) increases, a multivariate Gaussian tends towards the uniform distribution on that hypersphere. In addition, any two samples from \(\mathcal{N}\left(0,I_{n}\right)\) are always almost orthogonal to each other (this is called almost-orthogonality). A formal description of these phenomena can be found in~\cite{Vershynin2018High-DimensionalProbability}.

These facts are closely related to how \textit{(hyper-)volume} behaves in HD spaces: if we consider the standard uniform measure on the hypersphere, then most of its volume or mass is \textit{concentrated} in very thin `\textit{equatorial}' bands for \textit{any} randomly chosen north pole (this is, of course, just an intuitive statement, for a formal description see \cite{Wainwright2019ConcentrationMeasure}). The contrast with our intuition coming from two dimensional spheres is remarkable. In the next section, we will review how this may impact generative models like VAEs. 

\subsection{High Dimensional Spaces in Generative Models (VAE)}

One use case of the VAE is to generate data. Since the latent distribution is known (high dimensional multivariate Gaussian), one could sample from that distribution and decode the latent vector to generate a novel data sample. This works very well for simple data and low dimension latent spaces (e.g., MNIST with $n=2$, as done in \cite{DiederikPKingma2014Auto-EncodingBayes}), but not so well for more complicated data or high dimensional latent (e.g., CIFAR10, or MNIST with $n=32$ as in~\cite{Cinelli2021VariationalNetworks}). Indeed, variations of the VAE are commonly used in generative models of images and videos~\cite{JonathanHoAjayJain2020DenoisingModels}, but the latent space is not sampled directly: instead, a reverse diffusion process dynamically transforms a random sample into a valid latent location.

One of the reasons why VAE performs poorly when sampling directly from a high dimensional latent is because of conflicting constraints. On the one hand, high resolution images need many latent dimensions to capture all the information they convey. Then, the VAE's KL divergence term in the loss encourages the system to distribute this HD latent uniformly on the hypersphere. By doing so, the latent becomes extremely sparse given the number of training samples and the expanse of the latent hyperspherical manifold (the volume growing exponentially with the number of dimensions; in these regimes, trying to find a \textit{specific} latent point is akin to a `\textit{needle in the haystack}' kind of situation). The sparsity hampers any attempt to model the latent as a continuous manifold \cite{Peng2023InterpretingEffect}. But, on the other hand, this usually leads to meaningless decoding of random samples from the prior. These two opposing forces (sparsity due to HD spaces needed to model high resolution images vs need for a continuous manifold for meaningful generation from all of the latent space) combine to severely limit the capacity of VAEs to function as generative models. 

Since the samples in latent space live on the hypersphere, it comes naturally to consider using hyperspherical coordinates to describe latent variables.

\section{Related works}
\label{gen_inst}

 `Hyperspherical Variational Auto-Encoder' \cite{Davidson2018HypersphericalAuto-Encoders} proposed replacing the standard Euclidean KLD divergence with a KLD divergence between a uniform distribution on the hypersphere as a prior, and a von Mises-Fisher distribution as an approximate posterior. A different way of building a Hyperspherical Variational Auto-Encoder is proposed in \cite{Bonet2022SphericalSliced-Wasserstein}, based on a spherical Sliced-Wasserstein discrepancy, and as an extension of the well-known Euclidean models \cite{SoheilKolouriPhillipE.Pope2020Sliced-WassersteinModel}. 
Hyperspherical aspects of data are studied in \cite{Lowe2023RotatingDiscovery}, where high dimensional Rotating Features are introduced. Of particular interest for our project, the authors remarked: 

`\textit{We represent Rotating Features in Cartesian coordinates rather than (hyper) spherical coordinates, as the latter may contain singularities that hinder the model’s ability to learn good representations. [...]. This representation can lead to singularities due to dependencies between the coordinates. 
For example, when a vector’s radial component (i.e. magnitude) is zero, the angular coordinates can take any value without changing the underlying vector. As our network applies ReLU activation on the magnitudes, this singularity may occur regularly, hindering the network from training effectively}'. 

As we reviewed in Section I.B, in high dimensions, the random samples of an independent multivariate Gaussian distribution fall in the equator of a hypersphere, and thus none of them is near the singularities of the hyperspherical coordinates (the poles and the center of the hypersphere).

While the problem of formulating latent spaces given by non-Euclidean Riemannian manifolds, and hyperspheres in particular has been studied, explicit use of hyperspherical coordinates is avoided. At first look, the conversion from Cartesian to hyperspherical coordinates seems to require computationally expensive recurrent trigonometric formulas (see Appendix \ref{appendix:hstransform}). Instead, formulations of Riemannian geometry that rely on Cartesian coordinates are used. Riemannian geometry is not just about a curved metric, but also being able to express it in a convenient coordinate system \textit{adapted} to it (of central importance in physics).

The possible use of hyperspherical coordinates has thus been discarded by all the works that we reviewed. Notwithstanding the previous arguments, we believe that the use of hyperspherical coordinates is feasible and can be beneficial, as demonstrated by our numerical experiments. In Appendix \ref{appendix:hstransformcode} we provide a vectorized implementation for transforming between hyperspherical and Cartesian coordinates, which adds only a limited computational overhead for training a VAE.

Despite the fact that some of the models in the mentioned references also work on the hypersphere in latent space, the similitude ends there. We do not consider them suitable for comparison because both the goals and methods of these works are very different from ours. In particular, those approaches are not concerned with the sparsity issue of HD spaces as well as volume compression as a possible solution to it. \cite{Davidson2018HypersphericalAuto-Encoders}, for example (and related elaborations), build a VAE with a KLD term between a uniform distribution on the hypersphere and a von Mises-Fisher approximate posterior. Thus, by construction, there cannot be any compression of the type we discuss in our work, it is still highly similar to the standard VAE in the sense in which the posterior distribution ends approaching a uniform distribution on the hypersphere, where, we claim, the sparsity issue arises. 

In the next section, we formulate a VAE with the latent variables described using hyperspherical coordinates. 

\section{Method: VAE with hyperspherical coordinates}

Our approach is based on formulating the initial KL divergence term with a prior from the original VAE, which is in Cartesian coordinates, to one in hyperspherical coordinates. See Appendix \ref{appendix:hstransform} for the standard conversion formulas between Cartesian and hyperspherical in high dimension. 

In Cartesian coordinates, the KL divergence between the estimated posterior defined by $\mu_k$ and $\sigma_k$ and the prior defined by $\mu_k^p$ and $\sigma_k^p$ can be written as: 

\begin{align}
\text{KLD}_{\text{CartCoords}}^{w/Prior} \simeq \sum_{k=1}^{n} \Bigg( 
    &\left( \mathbb{E}_{b}[\sigma_{k}] - \sigma_{k}^{p} \right)^{2} 
    + \sigma_{b}[\sigma_{k}]^{2} \nonumber \\
    & + \left(\mathbb{E}_{b}[\mu_{k}] - \mu_{k}^{p} \right)^{2} 
    + \sigma_{b}[\mu_{k}]^{2} \Bigg)
\end{align}
where \(\mathbb{E}_b\) and \(\sigma_b\) denote the batch statistics over mini batches of data of size \(N_b\).

So far, not much has been gained other than rewriting the KLD function (in Cartesian coordinates) in terms of the batch statistics. This rewriting was partly inspired by the construction in \cite{Bardes2021VICReg:Learning}, and will be useful for our next step. 

We now introduce hyperspherical coordinates in the KLD formulation. We start with the Cartesian coordinates \((\mu_{i},\sigma_{i})\), given by the encoder, and transform these to their hyperspherical counterparts \((\overset{\mu}{r},\overset{\mu}{\varphi_{k}};\overset{\sigma}{r},\overset{\sigma}{\varphi_{k}})\) with $r$ a scalar and $k$ the index of the $n-1$ spherical angles. 

The KLD-like objective becomes for the angles $\varphi_k$:

\begin{align}
&\text{KLD}_{\text{HSphCoords}}^{w/Prior}(\varphi_k)= \nonumber\\
&\sum_{k=1}^{n-1} \bigg( \alpha_{\sigma,k} \left(\mathbb{E}_{b}[\mathrm{cos}\,\overset{\sigma}{\varphi_{k}}] - a_{\sigma,k} \right)^{2} +\beta_{\sigma,k}\left(\sigma_{b}[\mathrm{cos}\,\overset{\sigma}{\varphi_{k}}]-b_{\sigma,k}\right)^{2} \nonumber\\
&+ \alpha_{\mu,k}\!\!\left(\mathbb{E}_{b}[\mathrm{cos}\,\overset{\mu}{\varphi_{k}}]-a_{\mu,k}\right)^{2}\!\!\!\!+\!\!\beta_{\mu,k}\left(\sigma_{b}[\mathrm{cos}\,\overset{\mu}{\varphi_{k}}]-b_{\mu,k}\right)^{2} \!\bigg)\!\!\!
\end{align}

and for the norm $r$:

\begin{equation}
\begin{aligned}
&\hspace*{-2em}\text{KLD}_{\text{HSphCoords}}^{w/Prior}(r)=\\
{} & \alpha_{\sigma,r}\left(\mathbb{E}_{b}[\overset{\sigma}{r}]-a_{\sigma,r}\right)^{2}+\beta_{\sigma,r}\left(\sigma_{b}[\overset{\sigma}{r}]-b_{\sigma,r}\right)^{2} \\
& +\alpha_{\mu,r}\left(\mathbb{E}_{b}[\overset{\mu}{r}]-a_{\mu,r}\right)^{2}+\beta_{\mu,r}\left(\sigma_{b}[\overset{\mu}{r}]-b_{\mu,r}\right)^{2},
\end{aligned}
\end{equation}

with the priors for the mean over the batch $a_{i,j}$, the standard deviation over the batch $b_{i,j}$, and the gains for each term $\alpha_{i,j},\,\beta_{i,j}$, for $i\in\{\sigma,\mu\}$ and $j\in\{1,...,n-1,r\}$

We use the cosines rather than the angles to avoid costly extra computations of the corresponding arccosines (Appendix \ref{appendix:hstransform}). The coordinate transformation is done using a vectorized implementation (code provided in Appendix \ref{appendix:hstransformcode}). The reparameterization trick is still done in the Cartesian coordinates representation. The coordinate transformation and the extra KLD terms add about 32\% computation time during training \textit{per epoch} (measured at: 200 samples per batch, $n=200$). For more dimensions the increase is higher. The final cost to be optimized weights the reconstruction term and KLD terms with an overall gain $\beta$ for similarity with the standard $\beta$VAE (\ref{eq:VAEcost}):

\begin{eqnarray}
\mathcal{L} &=& \text{MSE}(x,\hat{x}_z) \nonumber \\   
&&+ \beta \left( \text{KLD}_{\text{HSphCoords}}^{w/Prior}(\varphi_k) +  \text{KLD}_{\text{HSphCoords}}^{w/Prior}(r)  \right).\label{eq:HSloss}
\end{eqnarray}

\subsection{Volume Compression of the latent manifold}

We discussed previously that the standard VAE forces the latent samples to be uniformly distributed on the hypersphere, which results, in high dimensions, in the data being located within equators of the hypersphere where the volume is the greatest. A benefit of using hyperspherical coordinates is the ability to set a prior for the $\varphi_k$ that forces the latent samples away from the equator, thereby escaping these regions. This can be done for each angular coordinate, as all are uncorrelated with each other, by simply setting 

\begin{equation}
a_{\mu,k} \neq0,\,\forall k.
\end{equation}

By doing so, the samples can be moved to a zone with much less volume, thereby increasing the density of the latent, with the hope that random samples from that denser region will have better quality decoding because of the reduced sparsity. This can be seen more directly by analyzing the hypervolume element of the hypersphere in hyperspherical coordinates. The volume can be reduced much faster and effectively by reducing the angular coordinates (away from the equators), than by either reducing just the radius of the hypersphere or, equivalently, all of the Cartesian coordinates. 

The higher the dimension, the more pronounced this difference becomes because each added dimension $k$ adds extra powers of $\sin\varphi_k$ in the hypervolume element (see Appendix \ref{appendix:hstransform}). Then, the further the angles from \(\pi/2\), the smaller the infinitesimal hypervolume element becomes as it is multiplied by an increasingly smaller quantity lower than $1$. This is a purely geometric effect. It can already be easily seen in the two-dimensional sphere, where a spherical coordinate rectangle of unvarying angular coordinates size has a smaller area when moved away from the equator towards any of the poles. Thus, high dimensions bring the problem of high volume in the equators, but also a non-Euclideanity to the manifold; we explored to which extent one can take advantage of the latter to mitigate the former.

Finally, by setting 

\begin{equation}
a_{\mu,r}=\sqrt{n}
\end{equation}

(and normalizing \(z\), after sampling via the reparameterization trick, to the same radius $\sqrt{n}$) we can force the latent samples to be on the hyperspherical surface of that radius. 

\section{Experimental Results}

\subsection{Model and implementation}

For all our experiments, we use a ResNet-type architecture \cite{He2015DeepRecognition} for both encoder and decoder. When using the loss in hyperspherical coordinates (\ref{eq:HSloss}), we use an annealing schedule \cite{Fu2019CyclicalVanishing} for the gain $\beta$ of the KLD-like loss, consisting of an initial stage which increases proportionally with \(\sqrt{\text{epoch}}\) for the first \(100\) epochs, and is constant afterwards. This was necessary because we observed that too much compression of the volume was detrimental to the performance, while a strong compression was still necessary at the initial stage. The total training was \(300\) epochs in all cases. 

\subsection{Choosing the gain for each loss terms}

The constants $\alpha_{i,j},\,\beta_{i,j}$ multiplying the elements of the hyperspherical loss are proportional to $1/\sqrt{k+1}$, where \(k\) is the coordinate index. This was necessary because, unlike the Cartesian coordinates, the hyperspherical coordinates are asymmetric and vary with $k$. This can be seen in the transformation formulas (Appendix \ref{appendix:hstransform}), where a product of an increasing amount of sine functions is necessary as the coordinate index increases. We chose $1/\sqrt{k+1}$, guided by the fact that the vector whose Cartesian coordinates are $(1,1,...,1)$ has a cosine of its spherical angles equal to $1/\sqrt{k+1}$ as the coordinate index, and because it gave the best results experimentally.

In this way, we were able to avoid lengthy calculations to obtain the mathematically exact formulas for both these constants and the KLD in hyperspherical coordinates, which we do not believe, anyway, to be of the most importance for the particular goals we had in this work.

\subsection{Visualisation of the latent in standard VAE}

A standard VAE with $128$ latent dimensions was trained using the MNIST dataset \cite{Y.LeCunetal1998Gradient-basedRecognition.}. Generating and decoding random samples from the prior latent resulted in meaningless decoded/generated data (Fig.\ref{fig:1}a), left panel). 

The latent hypersphere can be visualized in 3D as shown in Fig.\ref{fig:1}b), left. This was done by averaging the 128 latent dimensions into three (first 42, second 42, and the remaining 44), and normalizing each of the resulting 3D vectors to the sphere. Each latent vector could thus be plotted as a point in 3D, and shows a uniform-like distribution on the 2D sphere as expected. 

 This visualisation allows us to directly see the disorder and high multiplicity of the high volume regime. A k-NN classifier for the 10 classes of MNIST from the latent had an accuracy of \(0.95\), and 10 clusters can readily be seen when projecting the 128 latent dimensions into 2 (Fig.\ref{fig:1}c), left) using t-SNE. However, no particular clustering can be observed on the 3D visualisation (Fig.\ref{fig:1}b), left). Such a direct visualisation of the latent space \textit{cannot} display clusters, because they are buried into the `disorder' of the volume of the hypersphere equators, where most of the samples are located. There are many different possible points in latent space that can realize the same global configuration (the clustering). 

\subsection{Improved generation when the latent manifold is compressed}

Next, we train the same VAE using the same dataset but now with the KLD-type loss in hyperspherical coordinates (\ref{eq:HSloss}). The prior was set to \textit{compress} the (hyper-)volume in latent space by using $a_{\mu,k} =1,\,\forall k$, which pushes \textit{all} the $\varphi_k$ towards $0$. They could not be exactly $0$ because the reconstruction would all be the same. The reconstruction term balances the KLD term to spread the latent samples away from the angles $0$. 

\begin{figure*}[!h]
    \centering
    \includegraphics[width=1\linewidth]{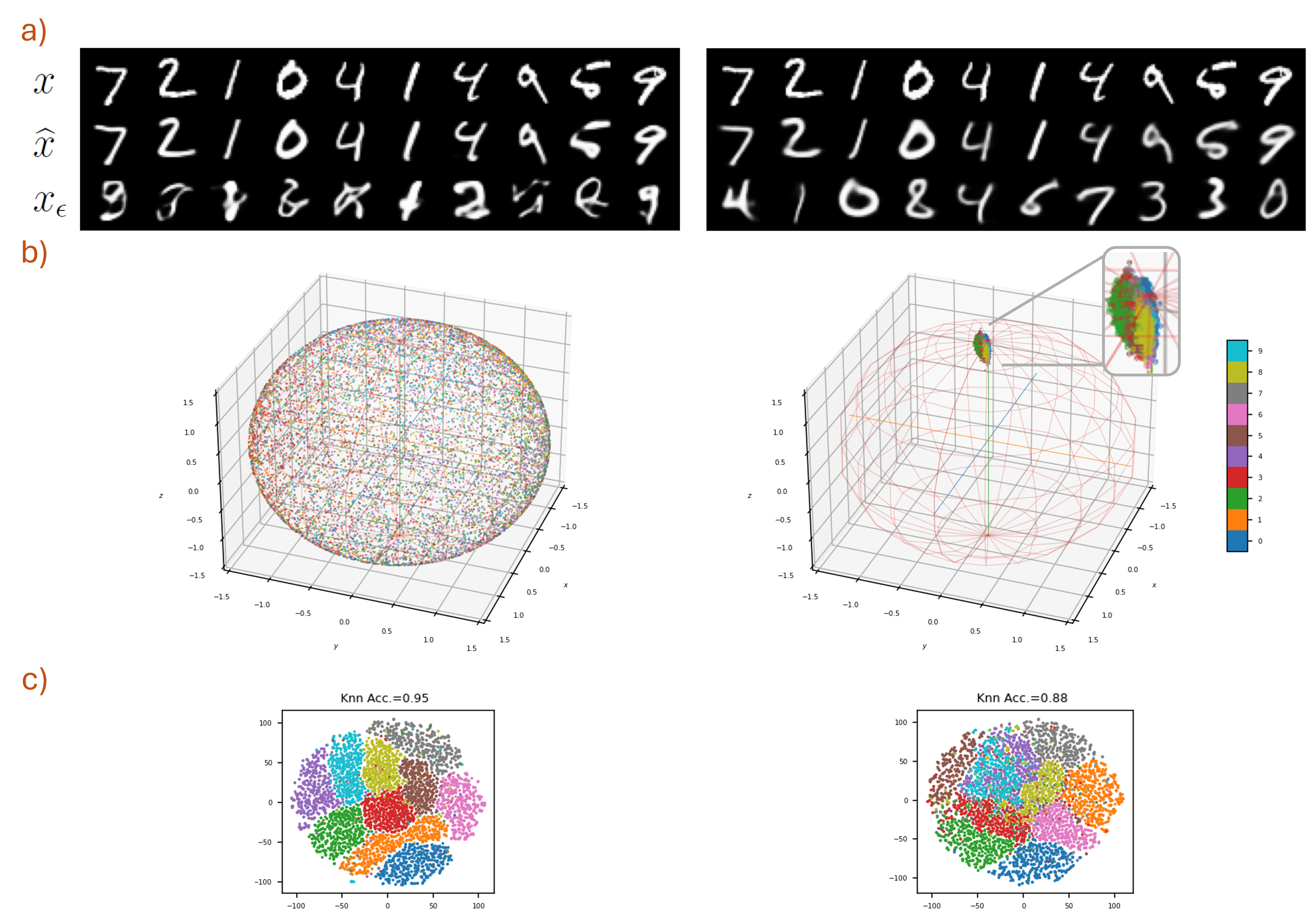}
    \caption{Comparison using MNIST between the standard $\beta$VAE (left) and the proposed compressed version (right). The top panel (a) shows the original data ($x$), the reconstruction ($\hat{x}$), and the generation sampling from the prior ($x_\epsilon$). The middle panel (b) shows the 3D projection on the latent 2D-sphere of the test dataset: the $\beta$VAE posterior is a uniform distribution with no visible clusters whereas the proposed method compresses the latent vector on a small volume within an island of the hypersphere and also makes the clusters visible. Despite this, the bottom panel (c) shows that in both cases the classes are clustered in the latent (using t-SNE) and that a k-NN classifier achieves good performance, with the compression $\beta$VAE resulting in lower accuracy (0.88 Vs 0.95) because the lower volume of the latent manifold forces the classes to overlap more (as seen on the clustering of panel c).}
    \label{fig:1}
\end{figure*}

In this configuration, for generating new data the latent was not randomly sampled on the whole hypersphere, but from a von Mises–Fisher distribution with the same mean and covariance as the ones empirically calculated from the latent embedding of the full test dataset. These decoded random samples generated data with a quality close to the actual training dataset (Fig.\ref{fig:1}a), right panel), to be compared to the meaningless decoding of the previous experiment when random sampling from the prior was done (Fig.\ref{fig:1}a), left panel). By compressing the latent using hyperspherical coordinates, the VAE became a functional generative model, despite having 128 latent variables.  

Furthermore, the \(3-\)dimensional visualization shows something remarkable (Fig.\ref{fig:1}b), right): besides showing that the latent samples are compressed towards a small `island' on the hypersphere and away from the equator, the classes are actually \textit{visible}. We believe that it is because the samples are now located away from the equator, in a region with a much lower volume, where there are many fewer possibilities to realize this clustering in terms of different possible latent point configurations. The same k-NN classifier from the latent space shows a similar accuracy, of  \(0.88\), and the class clusters can also be seen in a t-SNE 2D projection (Fig.\ref{fig:1}c), left).

\subsection{Trade off between reconstruction and generation}

The reconstruction quality of a VAE improves as the number of latent dimensions increases, as measured by the MSE between $x$ and $\hat{x}$. However, the quality of data generation (from decoding random sampling of the latent) decreases as the number of latent dimensions increases. We have argued in this paper that the later is due to the increased sparsity of the latent, as demonstrated qualitatively in the previous experiment when that sparsity is reduced by compressing the latent using our proposed method.

The quality of randomly generated data can be measured using the Frechet Inception Distance (FID) \cite{Heusel2017GANsEquilibrium}. FID compares the distribution of features between the images of the training/testing dataset and an equivalent number of randomly generated images.  We used in this experiment CIFAR10 \cite{Krizhevsky2009LearningImages}, a more challenging dataset, and an FID computed using 10,000 samples (we compare the random decoded samples with the \textit{reconstructed} testing set). We call this way of measuring the generation as `self-FID'.

In a VAE, the quality of the reconstruction, still measured by the MSE, also varies with the gain $\beta$ of the loss: the more weight for the KLD term, the more the latent matches the prior and the worse the reconstruction (Cf. $\beta$VAE \cite{Higgins2017-VAE:FRAMEWORK}).   

We can now explore quantitatively the quality of the reconstruction (using MSE) and the quality of the generation (using self-FID) when the number of latent dimensions increases and the $\beta$ varies. We compared the standard VAE with our proposed compressed VAE.

\begin{figure*}[!h]
    \centering
    \includegraphics[width=1.0\linewidth]{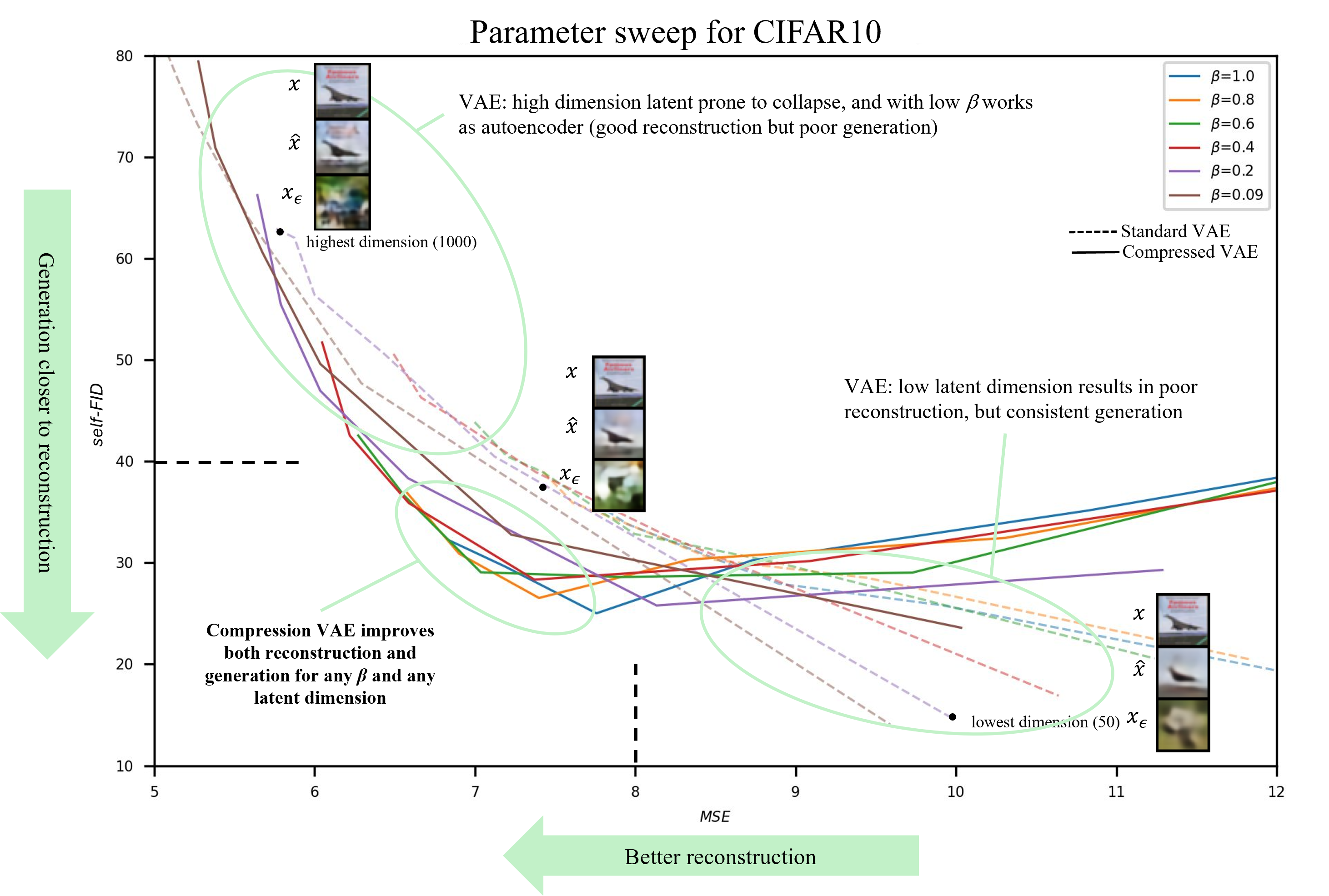}
    \caption{Effect of latent dimension and $\beta$ on the trade off between reconstruction and generation on CIFAR10. Each curve represents a VAE for a given $\beta$ while spanning the number of latent dimensions from low (50, bottom right endpoints) to high (1000 top left corner endpoints). The standard VAEs are shown using dashed lines, whereas the compressed versions are shown using solid lines. We excluded from our discussion the regimes where generation was of very poor quality (self-FID$>40$) or the reconstruction was too blurry (MSE$>8$), with the best trade off close to the bottom left corner. In that useful area, the compressed VAEs outperformed their standard equivalent for any combination of $\beta$ and latent size (solid lines closer to the bottom left corner than the dashed lines).}
    \label{fig:2}
\end{figure*}

These two metrics (MSE and self-FID), should give us a good idea regarding how good our models are for the general generative task: the fist measures how sharp (i.e., not `blurry') the reconstructed images are, while the second how close the random decoded images resemble images from the (reconstructed) training dataset. \textit{A good VAE-based generative model should minimize both of these metrics \textbf{simultaneously}: that is, to be able to generate random samples which are \textbf{in-distribution} w.r.t. the reconstructed dataset (low self-FID), and such that the latter actually resembles the original dataset (low MSE)}.

Fig.\ref{fig:2} sumarizes the results. As expected, the MSE decreases as one increases the latent space dimension, while the exact opposite is true for the self-FID. To obtain a good generative model in the way we defined it can be a very difficult task, involving a very delicate balance between these two opposing trends we just described, and often relying on off-equilibrium configurations.

Our experiments demonstrate that the compression VAE version improves on absolute terms over the standard VAE over any combination of $\beta$ and dimension of the latent. In Appendix \ref{appendix:celebaresults} we show similar results obtained with the CelebA dataset.

\section{Conclusion}

We propose to convert the latent variables of a VAE to hyperspherical coordinates. This allows moving latent vectors on a small island of the hypersphere, reducing sparsity. We showed that this modification improves the generation quality of a VAE. 
The following points will require further attention in regards to the present work:
\begin{itemize}
    \item the improvement in generation was only evaluated for the purposes of hypothesis testing, and not as absolute performance. Furthermore, the FID metric that we used may also have limitations sometimes \cite{Stein2023ExposingModels}.

    \item we did not evaluate the method for high resolution and larger datasets such as Imagenet.

    \item the extra computing time is about 32 per cent more per epoch for 200 latent dimensions. In (much) higher dimensions, the added computation increases and might become prohibitive.

    \item future research can focus in optimizing this method (or other method that takes into account the hypothesis about sparsity) for obtaining state-of-the-art results in generation and other tasks, in VAEs and other models.

    \item the use of latent representations in hyperspherical coordinates can also be further explored in several other applications (perhaps unrelated to compression and generation), by the use of the provided script for the conversion and inspired by its proof of concept of practical feasibility in the present paper.
\end{itemize}
\appendix
\section{Appendix}

\subsection{Conversion between Cartesian and hyperspherical coordinates}\label{appendix:hstransform}

For reference, we provide the standard formulas for converting between Cartesian and spherical coordinates.

In $n$ dimensions, given a set of Cartesian coordinates $x_k$ with $k\in\{1,\hdots,n\}$, the hyperspherical coordinates are defined by a radius $r$ and $n-1$ angles $\varphi_k$ with $k\in\{1,\hdots,n-1\}$; $\varphi_k \in [0,\hdots,\pi]$ for $k\in\{1,\hdots,n-2\}$ and $\varphi_{n-1} \in [0,\hdots,2\pi)$. 

From hyperspherical to Cartesian conversion:

\begin{equation}
\begin{aligned}
x_1 = & {} r \cos(\varphi_1) \\
x_2 = & {} r \sin(\varphi_1)\cos(\varphi_2) \\
x_2 = & {} r \sin(\varphi_1)\sin(\varphi_2)\cos(\varphi_3) \\
\vdots \\
x_{n-1} = & {} r \sin(\varphi_1)\sin(\varphi_2) \hdots \sin(\varphi_{n-2})\cos(\varphi_{n-1}) \\
x_{n} = & {} r \sin(\varphi_1)\sin(\varphi_2) \hdots \sin(\varphi_{n-2})\sin(\varphi_{n-1}) 
\end{aligned}
\end{equation}

From Cartesian to hyperspherical conversion:

\begin{equation}
\begin{aligned}
r = & {} \sqrt{x_{n}^2+x_{n-1}^2+\hdots+x_2^2+x_1^2} \\
\cos(\varphi_{1}) = & {} \frac{x_1}{ \sqrt{x_{n}^2+x_{n-1}^2+\hdots+x_2^2+x_1^2}}\\
\cos(\varphi_{2}) = & {} \frac{x_2}{ \sqrt{x_{n}^2+x_{n-1}^2+\hdots+x_2^2}}\\
\vdots \\
\cos(\varphi_{n-2}) = & {} \frac{x_{n-2}}{ \sqrt{x_{n}^2+x_{n-1}^2+x_{n-2}^2}}\\
\cos(\varphi_{n-1}) = & {} \frac{x_{n-1}}{ \sqrt{x_{n}^2+x_{n-1}^2}}\\
\end{aligned}
\end{equation}

The hypervolume element of the hypersphere $\mathbb{S}_{R}^{n-1}$ is given by the following expression when using hyperspherical coordinates\footnote{(see \url{https://en.wikipedia.org/wiki/N-sphere})}:

\begin{equation}
\begin{aligned}
\mathrm{d}V_{\mathbb{S}_{R}^{n-1}}=R^{n-1}\sin^{n-2}(\varphi_{1})\sin^{n-3}(\varphi_{2})\cdots \\\sin(\varphi_{n-2})\mathrm{d}\varphi_{1}\mathrm{d}\varphi_{2}\cdots\mathrm{d}\varphi_{n-1}
\end{aligned}
\end{equation}

\subsection{Vectorized code for converting between Cartesian and hyperspherical coordinates}\label{appendix:hstransformcode}

This code is accessible here and provided below for reference. 

\begin{lstlisting}

import torch

def cart_to_cos_sph (x, device):
    m = x.size(0)
    n = x.size(1)
    mask = torch.triu(torch.ones(n, n)).to(device)
    mask = torch.unsqueeze(mask, dim=0)
    mask = mask.expand(m, n, n)
    X = torch.unsqueeze(x, dim=1).expand(m, n, n)
    X_squared = torch.square(X)
    X_squared_masked = X_squared * mask
    denom = torch.sqrt(torch.sum(X_squared_masked, dim=2)+0.001)
    cos_phi = x / denom
    return cos_phi[:, 0:n-1]
\end{lstlisting}

\subsection{Results on CelebA64}\label{appendix:celebaresults}

In this appendix we include additional experimental results conducted on the dataset CelebA \cite{Liu2015DeepWild}, resized to a $64\times64$ image size.

The analysis is of the same type as the one we performed on CIFAR10 (cf. Fig.\ref{fig:2}).

\begin{figure*}[!h]
    \centering
    \includegraphics[width=1\linewidth]{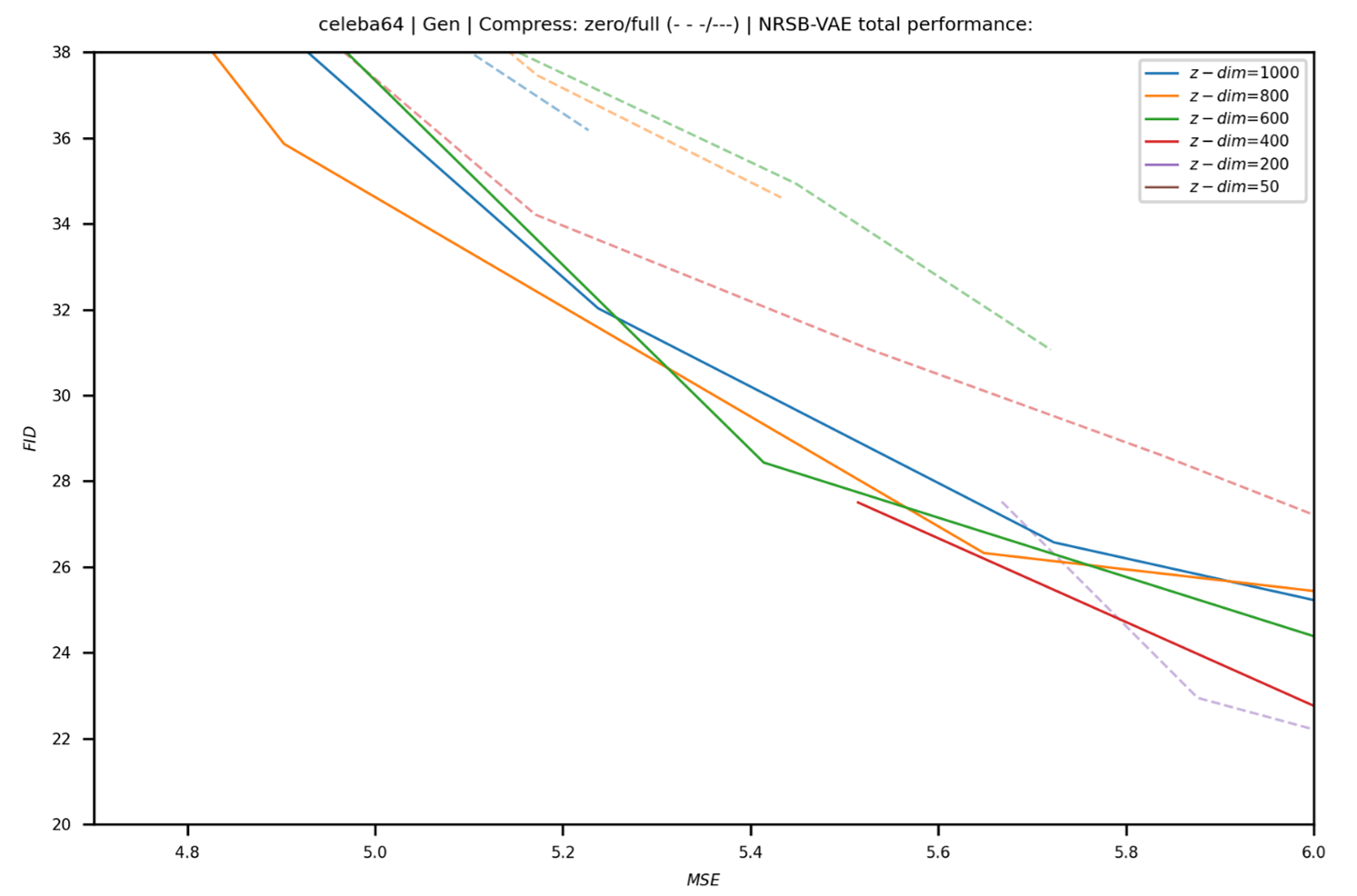}
    \caption{Effect of latent dimension and $\beta$ on the trade-off between reconstruction and generation on CelebA64 (as in CIFAR10, solid lines closer to the bottom left corner than the dashed lines).}
    \label{fig:18}
\end{figure*}

\bibliographystyle{IEEEtran}
\bibliography{paper_draft}

\end{document}